\documentclass[runningheads]{llncs}

\usepackage[T1]{fontenc}
\usepackage[utf8]{inputenc}
\usepackage{graphicx}
\usepackage{amsmath}
\usepackage{amssymb}
\usepackage{booktabs}
\usepackage{tabularx}
\usepackage{array}
\usepackage{fvextra}
\usepackage{placeins}
\usepackage{url}
\usepackage[hidelinks]{hyperref}

\title{Lost in Visual Translation: A VLM-Assisted Perceptual-Semantic Coherence Framework for EEG-to-Image Reconstruction}
\titlerunning{VLM-Assisted EEG-to-Image Evaluation}
\author{
Sukriti Tiwari\inst{1}\orcidID{0009-0008-9988-6492}, \\
BHVSP Subrahmanyam\inst{2}\orcidID{0009-0009-2759-3380}, \\
Nidhi Goyal\inst{1}\orcidID{0000-0001-8346-5951} and
Sai Amrit Patnaik\inst{3}\orcidID{0009-0000-5106-3210}
}

\authorrunning{S. Tiwari et al.}

\institute{
Department of Computer Science and Engineering, Mahindra University, Hyderabad, India\\
\email{\{sukriti.tiwari,nidhi.goyal\}@mahindrauniversity.edu.in}
\and
MU-VT Interdisciplinary Advanced Research Centre for Transformative Technologies, Mahindra University, Hyderabad, India\\
\email{subrahmanyam.bh@mahindrauniversity.edu.in}
\and
Avyakt Ehsaas\\
\email{saiamritp@gmail.com}
}

\begin{document}

\maketitle

\begin{abstract}
EEG-to-image evaluation should distinguish visual fidelity from recoverable meaning. Yet EEG-derived reconstructions are blurry, distorted, and low-detail, causing SSIM, LPIPS, and CLIP to penalize semantically recoverable outputs or reward plausible but incorrect ones. We analyze \textbf{6,855} ground-truth/reconstruction pairs from ATM, ENIGMA, BrainVis, and DreamDiffusion using semantic probes, caption harshness and blind-spot rates, and controlled degradations. Pixel metrics show near-zero correlation with semantic consistency, while representation metrics conflate perceptual and semantic errors. We therefore introduce a BCI-aware framework in which four VLMs assess image pairs through structured questions, producing Tolerant Perceptual Alignment Scores (T-PAS) and Tolerant Semantic Alignment Scores (T-SAS). Their consensus is distilled into the BCI-Coherence Score (BCS), a compact evaluator achieving T-PAS MAE \textbf{0.079} ($r=0.700$) and T-SAS MAE \textbf{0.082} ($r=0.850$) on our data. Human validation shows highly reliable joint coherence judgments ($\kappa=\textbf{0.882}\pm\textbf{0.174}$, $\alpha=\textbf{0.882}$), supporting perceptual--semantic recoverability over generic visual similarity. Code and resources are available at \url{https://sukt03.github.io/BCS/}.

\end{abstract}

\section{Introduction}

EEG-to-image reconstruction is an emerging neural decoding problem that aims to recover visual stimuli from non-invasive brain signals. Unlike conventional image restoration or image generation, the input in this setting is indirect, noisy, and spatially coarse. Consequently, even successful EEG-based reconstructions are often blurry, distorted or only partially faithful to the original stimulus. A reconstruction may fail to match the reference image at the level of pixels, texture, or fine visual detail, while still preserving enough structure or meaning for the original stimulus to remain recognizable.

\begin{figure}[t]
    \centering
    \includegraphics[width=\linewidth]{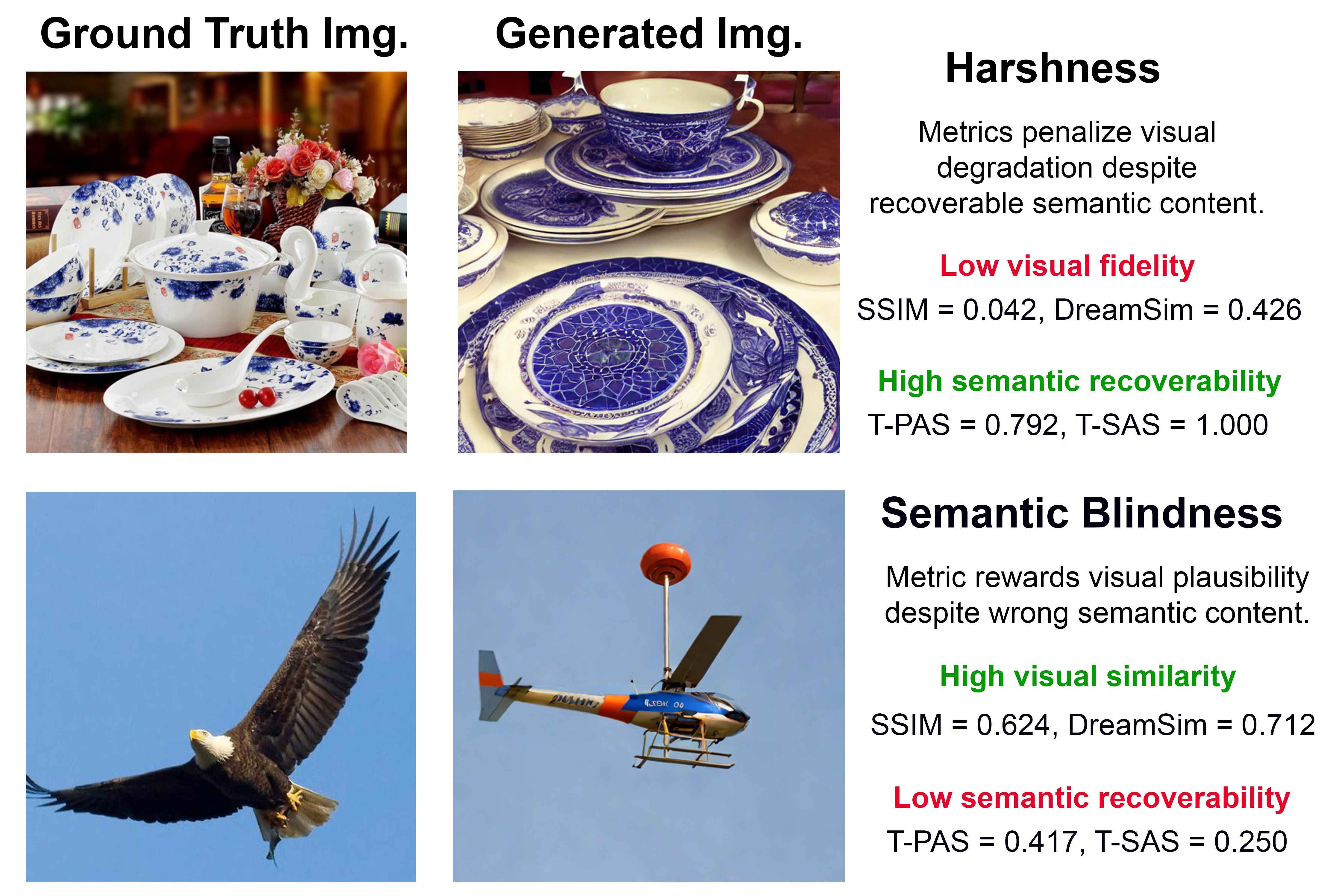}
    \caption{
 Representative failure patterns in EEG-to-image 
evaluation: harshness (top) and semantic blindness (bottom).
    }
    \label{fig:failure_patterns}
\end{figure}

Current evaluation practice in EEG-to-image reconstruction, however, largely inherits metrics from generic image reconstruction, perceptual similarity, and image generation. Commonly used measures include SSIM, PSNR, LPIPS, DISTS, DreamSim, CLIP/DINO-based similarity, and distributional or retrieval-style scores~\cite{wang2004ssim,ding2022dists,fu2023dreamsim,radford2021clip,oquab2023dinov2}. These metrics are useful in their intended settings, but they were not designed for the weak-signal, degradation-heavy regime of EEG decoding. Pixel and feature-level scores can penalize reconstructions for low visual fidelity even when the intended object or scene remains inferable, while semantic or embedding-based scores can reward broad similarity without verifying whether the reconstruction preserves the specific content of the reference stimulus. Prior neural decoding work has also shown that automatic reconstruction scores can diverge from human judgments~\cite{shen2019deep,takagi2023high,vanrullen2021natural}, yet recent EEG-to-image studies continue to rely on similar evaluation protocols despite the lower spatial fidelity and greater ambiguity of EEG compared with modalities such as fMRI~\cite{fei2024image,cao2025eegclip}.

This mismatch produces two complementary failure patterns. The first is \emph{harshness}: a metric may assign a low score to a reconstruction that is visually degraded but still semantically recoverable. In neural decoding, such an output may still carry meaningful information if the intended object, scene, or visual concept can be inferred despite imperfect rendering. The second is \emph{semantic blindness}: a reconstruction may appear perceptually plausible while depicting the wrong object category, scene, or visual content. Similar issues have been observed in image-generation evaluation, where semantic faithfulness often requires explicit object-, relation-, or question-based assessment rather than a single holistic similarity score~\cite{hu2023tifa,ghosh2023geneval,lin2024vqascore}. Figure~\ref{fig:failure_patterns} illustrates these two cases: harshness, where visual metrics penalize a semantically recoverable reconstruction, and semantic blindness, where a visually plausible reconstruction depicts incorrect semantic content. These observations suggest that EEG-to-image evaluation should not ask only whether a reconstruction is visually similar to the reference, but whether the reference stimulus remains perceptually and semantically recoverable under neural decoding noise.

\begin{figure}[t]
    \centering
    \includegraphics[width=\columnwidth]{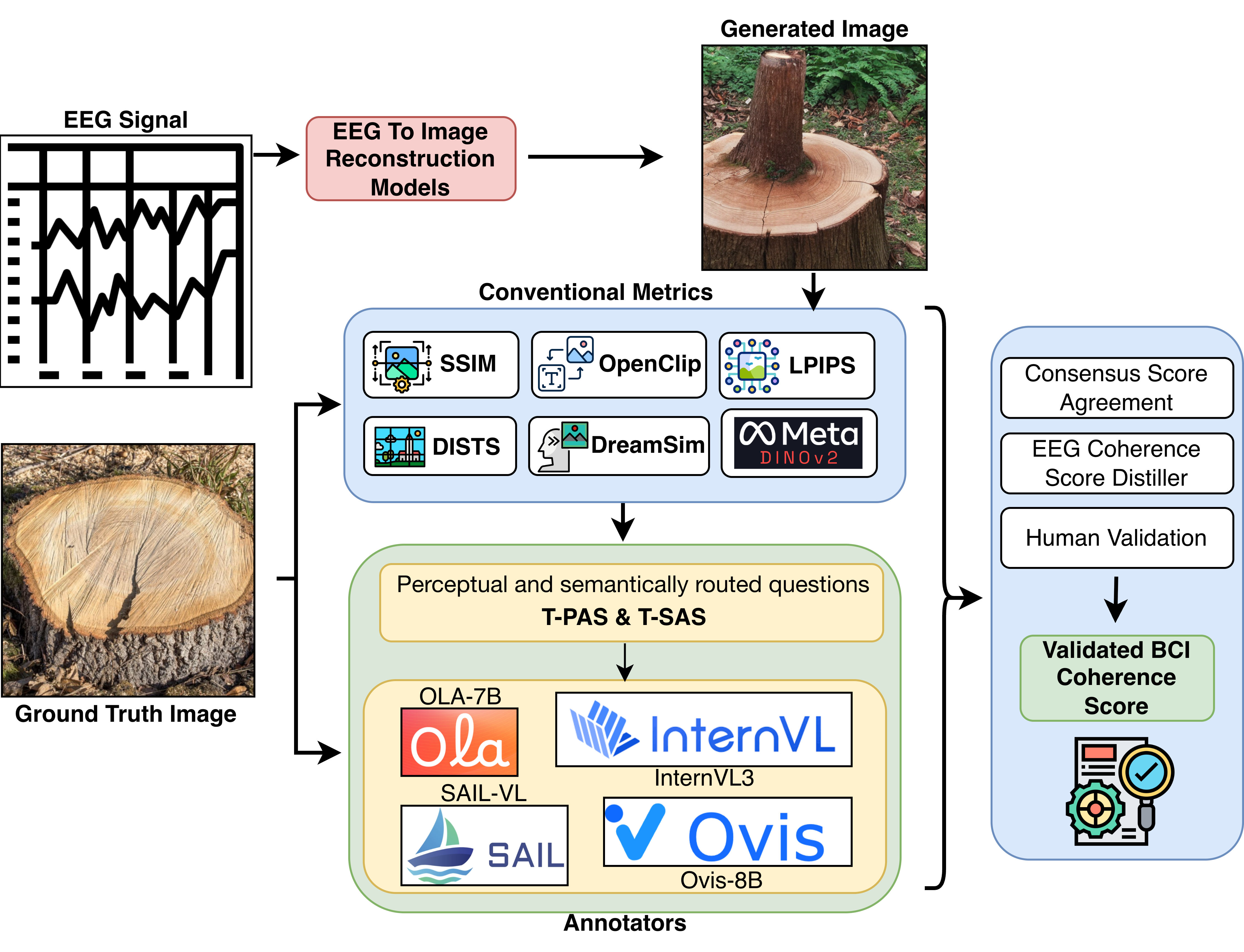}
    \caption{Overview of the proposed BCI-aware coherence evaluation framework.}
    \label{fig:bci}
\end{figure}

Motivated by this gap, we introduce a BCI-aware evaluation framework 
that separates visual degradation from semantic loss and measures how 
well a reconstruction preserves recoverable form, meaning, and their 
agreement. We use multiple VLMs as structured annotators, analyze their 
agreement, distill their consensus into a lightweight image-pair 
evaluator, and validate the resulting score with targeted human judgments 
on cases where perceptual and semantic assessments disagree. 
Figure~\ref{fig:bci} summarizes the proposed evaluation pipeline, 
including reconstruction outputs, conventional metric analysis, multi-VLM 
annotation, consensus aggregation, and BCS distillation.

Our contributions are as follows:

\begin{itemize}
\item \textbf{A perceptual--semantic analysis of EEG reconstruction metrics.}
We analyze 6,855 ground-truth/reconstruction pairs from EEG-to-image reconstruction outputs spanning ATM, ENIGMA, BrainVis, and DreamDiffusion models. We identify two failure regimes for conventional metrics: \emph{harshness}, where metrics penalize reconstructions with recoverable semantic content, and \emph{blind spots}, where metrics assign favorable scores despite semantic mismatch.

\item \textbf{A BCI-aware VLM annotation protocol.}
We introduce a structured annotation framework tailored to the artifacts and ambiguity of EEG-to-image reconstructions. Each image pair is evaluated through paired perceptual and semantic questions, yielding T-PAS (Tolerant Perceptual Alignment Score) and T-SAS (Tolerant Semantic Alignment Score). The protocol separates low-level visual recoverability from semantic recoverability under neural decoding noise and uses multi-VLM agreement and rationale similarity to assess annotation reliability.

\item \textbf{BCI-Coherence Score.}
We introduce BCI-Coherence Score (BCS), a compact evaluator that distills multi-VLM perceptual--semantic annotations into an efficient image-pair scoring model. Given a reference image and an EEG-based reconstruction, BCS predicts BCI-aware consensus scores without repeated access to large VLMs, making the evaluation signal easier to reuse and scale. We evaluate BCS on held-out reconstruction pairs and targeted human validation cases, showing that it preserves perceptual--semantic disagreement patterns that are recognizable to human evaluators.
\end{itemize}

\section{Related Work}

\textbf{Neural image reconstruction from brain signals.}
Visual reconstruction has progressed from decoding hierarchical visual features and optimizing images from fMRI to generative approaches based on adversarial networks, diffusion priors, and multimodal representations~\cite{horikawa2017generic,shen2019deep,ren2021reconstructing,chen2023seeing,takagi2023high,ozcelik2023braindiffuser,scotti2023mindeye,scotti2024mindeye2}. EEG offers greater accessibility and temporal resolution but substantially lower spatial resolution, signal-to-noise ratio, and cross-subject consistency. Early EEG reconstruction systems combined learned neural representations with conditional generation~\cite{singh2023eeg2image,guenther2024lowdensity}, while recent methods align EEG with CLIP or multimodal latent spaces and condition diffusion models on the resulting representations~\cite{bai2024dreamdiffusion,fei2024image,li2024visual,fu2025brainvis,cao2025eegclip,qi2026brain}. Their evaluation nevertheless mixes classification, retrieval, distributional quality, pixel fidelity, and embedding similarity. These measures answer different questions and may be influenced by powerful generative priors rather than information reliably decoded from neural activity~\cite{mayo2024brainbits}.

\textbf{Metrics for reconstruction and perceptual similarity.}
Distributional measures such as FID and KID evaluate generated collections but do not determine whether an individual reconstruction preserves its corresponding stimulus~\cite{heusel2017fid,binkowski2018kid}. Full-reference measures range from pixel and structural comparisons such as PSNR and SSIM to learned perceptual metrics including LPIPS, DISTS, PieAPP, and DreamSim~\cite{wang2004ssim,zhang2018lpips,ding2022dists,prashnani2018pieapp,fu2023dreamsim}. Representation similarities based on CLIP and DINO provide greater semantic sensitivity~\cite{radford2021clip,oquab2023dinov2}, but may overlook reference-specific errors or conflate visual resemblance with conceptual agreement. CLIP-IQA and Q-Align further demonstrate that vision-language representations can support human-aligned visual-quality assessment~\cite{wang2023clipiqa}. EEG reconstruction, however, requires separating tolerable signal-induced degradation from changes that alter the recoverable stimulus identity or meaning.

\textbf{Aspect-based and model-assisted evaluation.}
Text-to-image evaluation has increasingly replaced single holistic scores with decomposed tests of objects, attributes, relations, and compositional faithfulness. TIFA, GenEval, T2I-CompBench, Davidsonian Scene Graph, and VQAScore formulate evaluation through structured questions or verifiable semantic propositions~\cite{lin2024vqascore}. VIEScore and VLM-based quality assessors provide explainable multidimensional judgments, while ImageReward, PickScore, and VisionPrefer distill human or model preferences into reusable evaluators~\cite{lin2024viescore,kirstain2023pickapic}. Recent analyses also show that metric rankings depend strongly on prompts, question construction, and human-rating protocols. These approaches primarily assume prompt-conditioned, visually polished outputs. In contrast, EEG reconstruction is reference-grounded and systematically degraded by neural uncertainty. We therefore introduce a degradation-tolerant protocol that separately measures perceptual recoverability, semantic recoverability, and their joint coherence through the BCI-Coherence Score~(BCS).

\section{Limitations of Current EEG Reconstruction Metrics}
\label{sec:metric_audit}

\label{sec:metrics} 

We examine whether commonly used EEG-to-image reconstruction metrics capture two properties needed for reliable evaluation: semantic consistency with the reference image and sensitivity to controlled perceptual degradation.

\subsection{Caption-Level Semantic Probe}

To obtain an external semantic reference signal, we caption each ground-truth image and reconstruction using BLIP and compute the SBERT cosine similarity between the two captions:
\begin{equation}
    c_i =
    \mathrm{SBERT}\big(
    \mathrm{BLIP}(y_i),
    \mathrm{BLIP}(\hat{y}_i)
    \big).
\end{equation}
Here, $y_i$ denotes the ground-truth image and $\hat{y}_i$ denotes the corresponding EEG-based reconstruction. The caption similarity score $c_i$ is used as a semantic probe rather than a complete reconstruction metric, since captions can miss fine-grained visual details. It nevertheless provides a useful test of whether conventional metrics preserve a basic semantic ordering.

For each conventional metric $m$, we orient its score so that larger values consistently indicate better reconstruction quality:
\begin{equation}
q_i^m =
\begin{cases}
m(y_i,\hat{y}_i), & \text{if higher values are better},\\
-m(y_i,\hat{y}_i), & \text{if lower values are better}.
\end{cases}
\end{equation}
This produces an oriented metric score $q_i^m$ for each image pair. We then compare the ranking induced by $q_i^m$ with the caption-level semantic consistency score $c_i$.

\subsection{Caption-Based Failure Rates}

We define two quartile-based failure rates that capture complementary forms of metric misalignment. The first is the \emph{caption harshness rate}, which measures how often a metric assigns a bottom-quartile score to reconstructions whose captions remain semantically close to the reference:
\begin{equation}
\mathrm{CaptionHR}(m) =
\frac{
\left|\{i : c_i \geq Q_{75}(c),\; q_i^m < Q_{25}(q^m)\}\right|
}{
\left|\{i : c_i \geq Q_{75}(c)\}\right|
}.
\end{equation}
A high value indicates that the metric penalizes reconstructions that preserve caption-level semantics, typically because it is sensitive to visual degradation, texture mismatch, or other low-level distortions.

The second is the \emph{caption blind-spot rate}, which measures how often a metric assigns a top-quartile score to reconstructions whose captions are semantically dissimilar to the reference:
\begin{equation}
\mathrm{CaptionBSR}(m) =
\frac{
\left|\{i : q_i^m \geq Q_{75}(q^m),\; c_i < Q_{25}(c)\}\right|
}{
\left|\{i : q_i^m \geq Q_{75}(q^m)\}\right|
}.
\end{equation}
A high value indicates that the metric can reward reconstructions even when the caption-level semantic content differs from the ground truth. For these quartile-based rates, values near $0.25$ correspond to behavior close to chance with respect to caption-level semantic consistency.

\begin{table}[t]
\centering
\caption{\footnotesize Caption-level semantic consistency of conventional reconstruction metrics. Caption-HR measures harshness; Caption-BSR measures blind spots. Lower is better for both.}
\label{tab:caption_metric_audit}
\small
\setlength{\tabcolsep}{4pt}
\renewcommand{\arraystretch}{1.05}
\begin{tabular}{@{}lcccc@{}}
\toprule
\textbf{Metric} &
$\rho_{\text{cap}}$ &
$r_{\text{cap}}$ &
\textbf{Cap-HR} $\downarrow$ &
\textbf{Cap-BSR} $\downarrow$ \\
\midrule
MSE         & -0.012 & -0.003 & 0.257 & 0.257 \\
SSIM        & -0.044 & -0.059 & 0.277 & 0.247 \\
LPIPS       &  0.109 &  0.097 & 0.211 & 0.188 \\
DISTS       &  0.229 &  0.238 & 0.167 & 0.159 \\
DreamSim    &  0.483 &  0.526 & 0.071 & 0.092 \\
OpenCLIP    &  0.501 &  0.553 & 0.060 & 0.082 \\
DINOv2      &  0.389 &  0.503 & 0.103 & 0.116 \\
ImageReward &  0.309 &  0.461 & 0.113 & 0.152 \\
\bottomrule
\end{tabular}
\end{table}

\subsection{Controlled Perceptual Degradation Probe}

We next test whether the same metrics track perceptual degradation when the corruption process is known. We construct variants of the ground-truth images using six degradations: Gaussian blur, additive noise, downsampling, JPEG compression, color attenuation, and spatial shift. Each degradation is applied at four increasing severity levels, allowing metric behavior to be evaluated without VLM scores or human annotations.

For each ground-truth image $y_i$, degradation type $d$, and severity level $k$, we generate a corrupted image $\tilde{y}_{i,k}^{(d)}$. For a metric $m$, we orient its score so that larger values indicate better quality:
\begin{equation}
q_{i,k}^{m,d} =
\begin{cases}
m\!\left(y_i,\tilde{y}_{i,k}^{(d)}\right), & \text{if higher values are better},\\
-m\!\left(y_i,\tilde{y}_{i,k}^{(d)}\right), & \text{if lower values are better}.
\end{cases}
\end{equation}
We then measure whether the oriented score decreases monotonically as degradation severity increases:
\begin{equation}
\rho_{\mathrm{deg}}(m,d)
=
\rho_{\mathrm{Spearman}}
\!\left(
q_{i,k}^{m,d}, -k
\right).
\end{equation}
A high value indicates that the metric consistently tracks increasing perceptual corruption, while a low value indicates weak sensitivity or instability for that degradation type.

\begin{table}[t]
\centering
\caption{\footnotesize Controlled perceptual degradation probe. Entries are Spearman correlations between oriented metric quality and negative degradation severity; higher is better.}
\label{tab:perceptual_degradation_probe}
\small
\setlength{\tabcolsep}{3.2pt}
\renewcommand{\arraystretch}{1.05}
\begin{tabular}{@{}lrrrrrrr@{}}
\toprule
\textbf{Metric} &
$n$ &
\textbf{Blur} &
\textbf{Noise} &
\textbf{Down.} &
\textbf{JPEG} &
\textbf{Color} &
\textbf{Shift} \\
\midrule
MSE         & 1085 & 0.738 & 0.968 & 0.779 & 0.612 &  0.625 & 0.605 \\
PSNR        & 1085 & 0.738 & 0.968 & 0.779 & 0.612 &  0.625 & 0.605 \\
SSIM        & 1085 & 0.720 & 0.914 & 0.844 & 0.717 &  0.579 & 0.233 \\
Edge cosine & 1085 & 0.947 & 0.905 & 0.960 & 0.836 & -0.607 & 0.062 \\
Color hist. & 1085 & 0.764 & 0.943 & 0.773 & 0.727 &  0.631 & 0.898 \\
CLIP-L/14   &   64 & 0.934 & 0.823 & 0.948 & 0.734 &  0.890 & 0.524 \\
SigLIP      &   64 & 0.915 & 0.890 & 0.927 & 0.701 &  0.877 & 0.489 \\
\bottomrule
\end{tabular}
\end{table}

\subsection{Results and Implications}

Table~\ref{tab:caption_metric_audit} shows that pixel-level metrics are weak semantic indicators in EEG-to-image reconstruction. MSE and SSIM are nearly uncorrelated with caption-level semantic consistency, with $\rho_{\text{cap}}=-0.012$ and $\rho_{\text{cap}}=-0.044$, respectively, and their Caption-HR and Caption-BSR values remain close to the chance quartile rate. Representation-based metrics such as OpenCLIP and DreamSim align better with caption semantics, but still compress object identity, scene context, shape, color, texture, and artifacts into a single scalar.

The controlled degradation probe in Table~\ref{tab:perceptual_degradation_probe} shows a complementary limitation: metric behavior depends strongly on the type of visual corruption. Pixel-level metrics track additive noise well but are less stable under JPEG compression, color attenuation, and spatial shift. Edge and color probes are useful for specific degradations but fail outside their intended attribute, while CLIP-L/14 and SigLIP are more robust globally but do not reveal which perceptual property has degraded.

Together, the semantic and perceptual probes show that no single metric
tracks both semantic consistency and perceptual degradation reliably:
metrics sensitive to low-level corruption are weak semantic indicators,
while representation metrics that better capture global similarity still
obscure specific perceptual and semantic failure modes. 

\section{BCI-Aware Scoring and Evaluation}
\label{sec:bci_scoring}

The analysis in Sec.~\ref{sec:metrics} motivates a protocol that 
separately measures perceptual and semantic recoverability. The goal 
is not to judge photorealism, but to determine whether the reference 
stimulus remains recoverable under the distortions typical of neural 
decoding.
\subsection{Annotation Protocol}
\label{sec:annotation_protocol}

For each reconstruction pair, the VLM is shown the ground-truth image $y_i$ and the generated image $\hat{y}_i$. The model is instructed to compare the two images under the assumption that $\hat{y}_i$ is a noisy EEG-based reconstruction. Each question is answered using a three-level scale:
\begin{equation}
    \phi(\texttt{no}) = 0,\qquad
    \phi(\texttt{somewhat}) = 0.5,\qquad
    \phi(\texttt{yes}) = 1.
\end{equation}
In addition to the discrete answer, the VLM provides a short one to two-line rationale. This rationale is not used directly in score computation, but is used to assess whether different VLMs justify their judgments using similar visual or semantic evidence.

The annotation questions are designed to separate visual recoverability from semantic recoverability. Perceptual questions assess whether the reconstruction preserves layout, shape, texture, color, artifact severity, and overall visual interpretability. Semantic questions assess whether the reconstruction preserves category identity, fine-grained identity, functional role, quantity, scene context, and overall meaning. Table~\ref{tab:vlm_questions} lists the exact questions used in the protocol; the full prompt specification is provided in Appendix Sect.~\ref{app:prompt}.

\begin{table*}[t]
\centering
\caption{BCI-aware VLM annotation questions. Each question is answered using \texttt{no}, \texttt{somewhat}, or \texttt{yes}.}
\label{tab:vlm_questions}
\small
\setlength{\tabcolsep}{4pt}
\renewcommand{\arraystretch}{1.08}
\begin{tabularx}{\textwidth}{@{}llX@{}}
\toprule
\textbf{Group} & \textbf{ID} & \textbf{Question} \\
\midrule
Perceptual & P1 & \textbf{Spatial layout:} Does the generated image preserve the overall layout and position of the main regions? \\
Perceptual & P2 & \textbf{Shape:} Is the main object's shape or silhouette similar? \\
Perceptual & P3 & \textbf{Texture/material:} Are surface texture or material patterns similar? \\
Perceptual & P4 & \textbf{Color:} Are the main colors and chromatic appearance similar? \\
Perceptual & P5 & \textbf{Artifacts:} Is the image free from severe artifacts or noise that obscure the content? \\
Perceptual & P6 & \textbf{Holistic visual recoverability:} Can the original visual content be broadly recovered from the generated image? \\
\midrule
Semantic & S1 & \textbf{Basic category:} Is the main object or scene category correct? \\
Semantic & S2 & \textbf{Specific identity:} Is the specific subtype or identity correct, rather than only the broad category? \\
Semantic & S3 & \textbf{Function/purpose:} Does the generated content preserve the object's role or purpose? \\
Semantic & S4 & \textbf{Quantity:} Is the number of main objects or entities preserved? \\
Semantic & S5 & \textbf{Scene/context:} Is the surrounding scene or environment similar? \\
Semantic & S6 & \textbf{Holistic semantic recoverability:} Can the original meaning or content be understood from the generated image? \\
\bottomrule
\end{tabularx}
\end{table*}

Each analyzed pair is scored with the same six perceptual and six semantic questions.

For each image pair, the perceptual aggregate score T-PAS and semantic aggregate score T-SAS are computed as
\begin{equation}
    \mathrm{T\text{-}PAS}_i^{(v)}
    =
    \frac{1}{6}
    \sum_{p=1}^{6}
    \phi(a_{i,p}^{(v)}),
\end{equation}
\begin{equation}
    \mathrm{T\text{-}SAS}_i^{(v)}
    =
    \frac{1}{6}
    \sum_{s=1}^{6}
    \phi(a_{i,s}^{(v)}),
\end{equation}
where $a_{i,p}^{(v)}$ and $a_{i,s}^{(v)}$ are the answers from VLM $v$ for perceptual and semantic questions, respectively.

For downstream consensus supervision, we aggregate across VLMs at the question level:
\begin{equation}
    \tilde{a}_{i,q}
    =
    \mathrm{median}_{v \in \mathcal{V}}
    \phi(a_{i,q}^{(v)}),
\end{equation}
where $\mathcal{V}$ is the set of VLM annotators. With four annotators, this median can take intermediate values such as $0.25$ or $0.75$, preserving partial disagreement between models rather than forcing a hard majority label.

\subsection{VLM Evaluation Setup}
\label{sec:four_vlm_setup}

We annotate all 6,855 image pairs using four recent VLMs: InternVL3, SAIL-VL, OLA-7B, and Ovis2-8B. We select these VLMs because they are open-weight, reproducible, and representative of strong contemporary VLM performance as demonstrated in Open VLM Leaderboard on Hugging Face. Their open availability allows the annotation protocol to be reproduced without relying on paid proprietary APIs, while their competitive performance on public VLM benchmarks makes them suitable choices for large-scale annotation. The models also differ in calibration and architecture, providing a useful basis for consensus scoring rather than treating any single VLM as an oracle. Each model is evaluated with the same question set and output schema. All final runs produced valid annotations for all image pairs; checkpoint and inference settings are listed in Appendix Sect.~\ref{app:vlm_annotators}.

\begin{table}[t]
\centering
\caption{Aggregate scores from the four VLM annotators. Scores are averaged over the analyzed image pairs.}
\label{tab:vlm_score_summary}
\small
\setlength{\tabcolsep}{8pt}
\renewcommand{\arraystretch}{1.08}
\begin{tabular}{@{}lrr@{}}
\toprule
\textbf{VLM} & \textbf{Mean T-PAS} & \textbf{Mean T-SAS} \\
\midrule
InternVL3 & 0.341 & 0.252 \\
SAIL-VL   & 0.213 & 0.157 \\
OLA-7B    & 0.433 & 0.335 \\
Ovis2-8B  & 0.337 & 0.345 \\
\bottomrule
\end{tabular}
\end{table}

Table~\ref{tab:vlm_score_summary} shows that the annotators have different operating points. SAIL-VL is the most conservative, assigning lower perceptual and semantic scores. OLA-7B is more permissive, especially perceptually. InternVL3 and Ovis2-8B are closer in perceptual score, while Ovis2-8B assigns the highest semantic aggregate. This variation motivates using a consensus signal rather than treating any single VLM as the sole authority.

\begin{figure*}[t]
    \centering
    \includegraphics[width=0.47\textwidth]{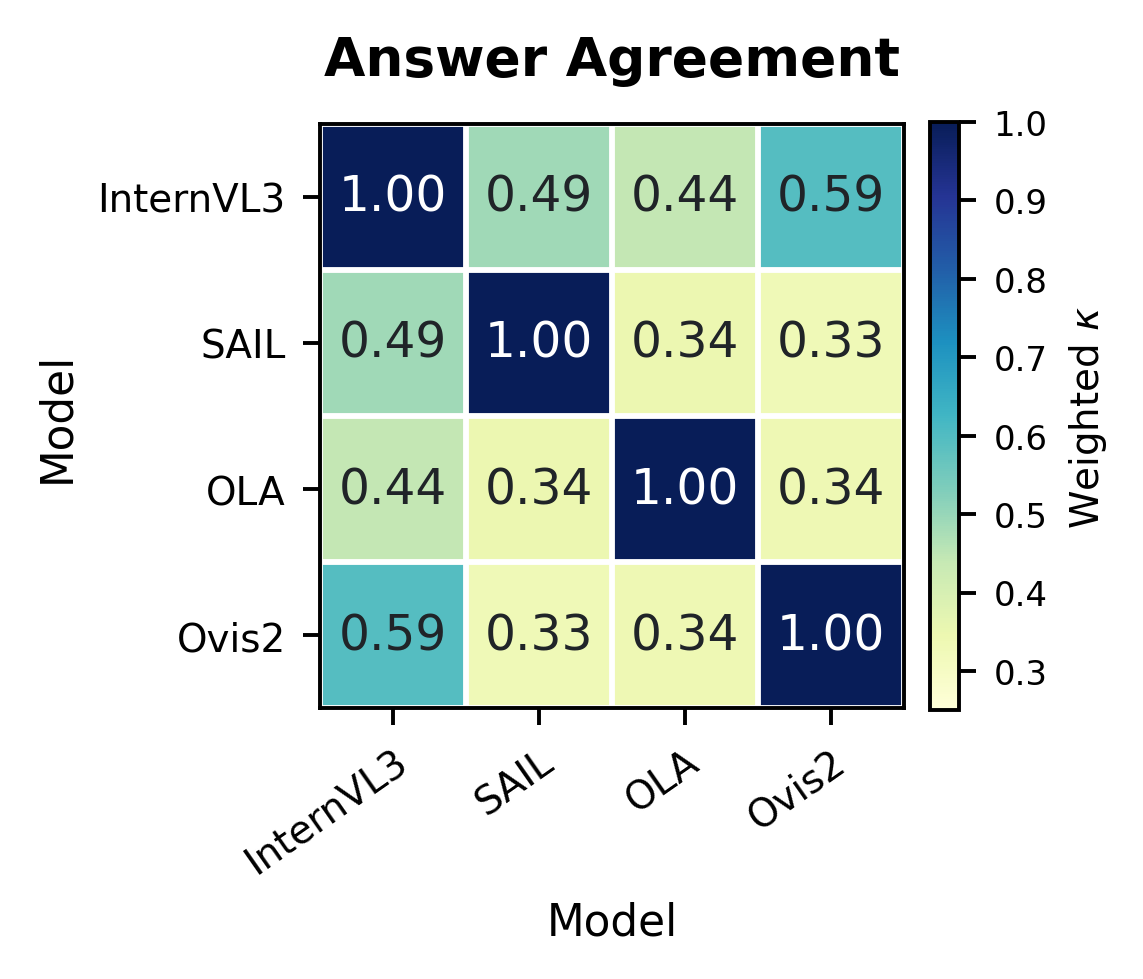}
    \hfill
    \includegraphics[width=0.47\textwidth]{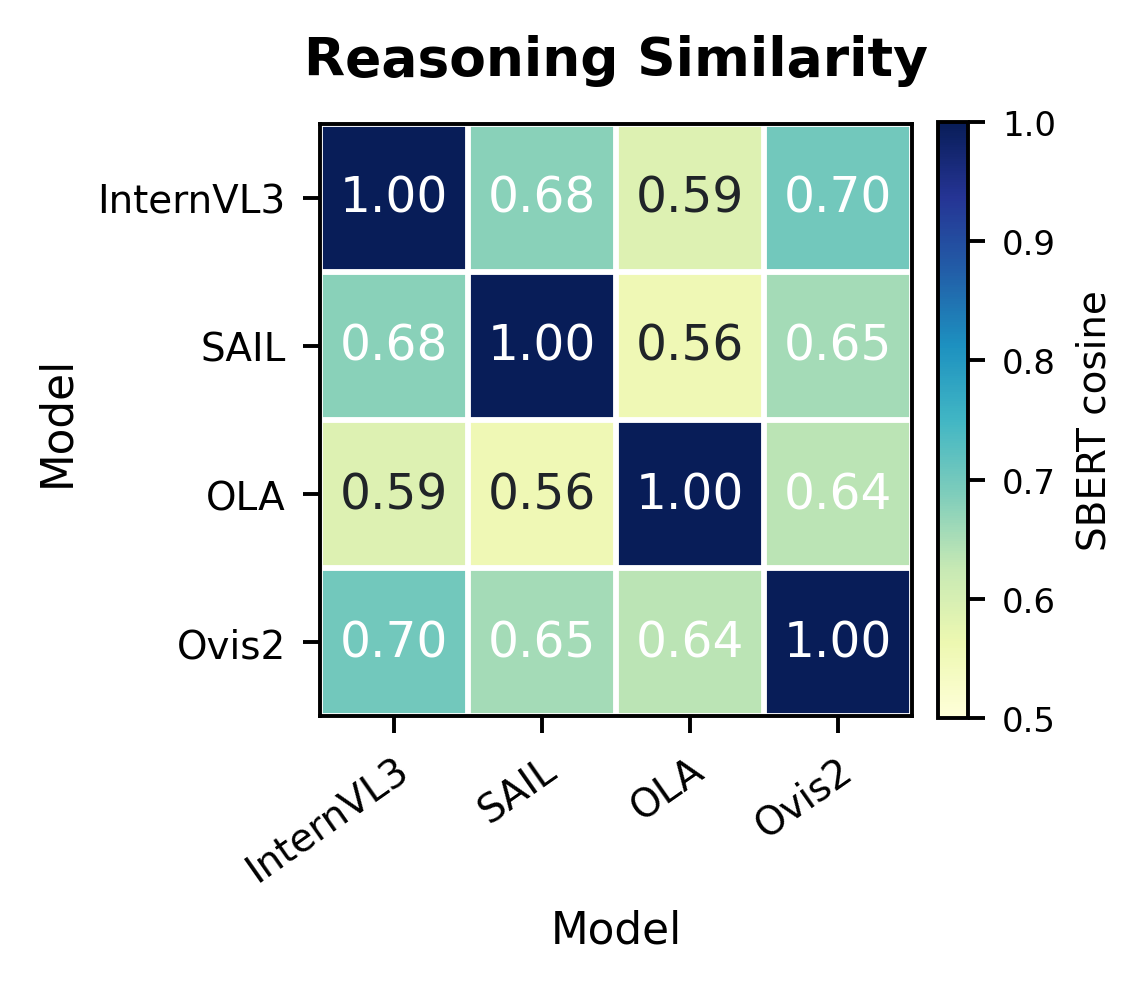}
    \caption{
    Pairwise VLM agreement and reasoning similarity. Left: answer agreement measured by weighted $\kappa$. Right: rationale similarity measured by SBERT cosine similarity. The matrices show that model pairs differ in answer calibration, while rationale similarity remains moderate to high across most pairs.
    }
    \label{fig:vlm_pairwise_matrices}
\end{figure*}

\subsection{Agreement and Reliability Analysis}
\label{sec:vlm_agreement}

We evaluate annotation reliability at two levels: answer agreement and reasoning similarity. Answer agreement is computed over all image-pair/question instances. Each image pair is scored with 12 questions: six perceptual questions and six semantic questions.

This gives 82,260 scored question instances in total. For each scored instance, we compare the four VLM answers. The unanimous agreement rate is 44.7\%, and Fleiss' $\kappa$ is 0.363. This indicates moderate agreement, which is expected because the task requires judging heavily degraded reconstructions rather than clean natural images.

Figure~\ref{fig:vlm_pairwise_matrices} shows the pairwise agreement structure across the four VLM annotators. InternVL3 and Ovis2-8B show the strongest agreement, with weighted $\kappa=0.594$, and also the highest rationale similarity of 0.700. SAIL-VL and OLA-7B differ more strongly in answer calibration, but the rationale-similarity matrix shows that model explanations remain moderately aligned across most pairs. This suggests that even when discrete answers differ, the VLMs often rely on related visual or semantic evidence.

We further inspect which dimensions are stable across models. Table~\ref{tab:question_agreement} reports representative low and high-agreement dimensions.

\begin{table}[t]
\centering
\caption{Question-level VLM agreement by annotation dimension. Unan. denotes unanimous answer agreement across the four VLMs.}
\label{tab:question_agreement}
\small
\setlength{\tabcolsep}{4pt}
\renewcommand{\arraystretch}{1.08}
\begin{tabularx}{\linewidth}{@{}Xlcc@{}}
\toprule
\textbf{Dimension} & \textbf{Type} & \textbf{Unan.} & \textbf{Fleiss' $\kappa$} \\
\midrule
P6 Holistic visual recoverability & Low  & 0.040 & -0.133 \\
P1 Global spatial structure       & Low  & 0.058 & -0.060 \\
S6 Semantic recoverability        & Low  & 0.071 &  0.012 \\
P5 Artifact absence               & Low  & 0.098 & -0.035 \\
\midrule
S2 Subordinate identity           & High & 0.769 & 0.289 \\
S5 Scene context                  & High & 0.784 & 0.383 \\
S3 Functional role/purpose        & High & 0.819 & 0.499 \\
P2 Object shape/silhouette        & High & 0.826 & 0.418 \\
\bottomrule
\end{tabularx}
\end{table}

This analysis reveals an important property of the protocol. Concrete attributes such as object shape, functional role, and scene context are relatively stable across VLMs. In contrast, holistic questions such as visual or semantic recoverability are more subjective because they require judging whether a degraded reconstruction is still interpretable. Rather than hiding this uncertainty, the multi-VLM setup exposes it and allows downstream consensus scores to reflect partial agreement.

\subsection{Discussion}
\label{sec:vlm_discussion}

Overall, the four-VLM analysis supports the use of consensus supervision. The models produce valid structured annotations at scale, differ meaningfully in calibration, and show higher agreement on concrete attributes than on holistic recoverability judgments. We therefore use the multi-VLM consensus signal for subsequent analysis and for training the BCI-Coherence Score. Additional metric-to-consensus correlations are reported in Appendix Sect.~\ref{app:metric_consensus}.

\section{BCI-Coherence Score}
\label{sec:bci_coherence}

The VLM annotation protocol in Sec.~\ref{sec:bci_scoring} provides detailed perceptual and semantic judgments, but running multiple VLMs for every new reconstruction set is computationally expensive. We therefore introduce the \emph{BCI-Coherence Score} (BCS), a compact VLM-distilled evaluator that predicts BCI-aware perceptual--semantic scores directly from a ground-truth image and its reconstruction.

\subsection{Distillation Target}
\label{sec:bcs_target}

For each image pair $(y_i,\hat{y}_i)$ and annotation question $q$, we aggregate the four VLM judgments into a consensus target. Let $a_{i,q}^{(v)} \in \{\texttt{no}, \texttt{somewhat}, \texttt{yes}\}$ denote the answer from VLM $v$, mapped to numerical values by
\begin{equation}
    \phi(\texttt{no}) = 0,\qquad
    \phi(\texttt{somewhat}) = 0.5,\qquad
    \phi(\texttt{yes}) = 1.
\end{equation}
The consensus target is the median score across VLMs:
\begin{equation}
    z_{i,q}
    =
    \mathrm{median}_{v \in \mathcal{V}}
    \phi(a_{i,q}^{(v)}).
\end{equation}
With four VLMs, this target can take intermediate values such as $0.25$ or $0.75$, preserving partial disagreement between annotators rather than forcing a hard majority label. We train BCS to predict the perceptual and semantic question scores:
\begin{equation}
    \hat{\mathbf{z}}_i
    =
    f_\psi(y_i,\hat{y}_i),
\end{equation}
where $f_\psi$ is a lightweight image-pair regression model.

\subsection{Model Architecture}
\label{sec:bcs_architecture}

BCS uses frozen visual encoders to extract image features from the ground-truth and reconstructed images. In our strongest variant, we use a fusion of SigLIP, CLIP, and DINOv3 image encoders. For each encoder $E_k$, we compute normalized embeddings
\begin{equation}
    \mathbf{e}_{i,k}^{\mathrm{gt}} = E_k(y_i),
    \qquad
    \mathbf{e}_{i,k}^{\mathrm{rec}} = E_k(\hat{y}_i).
\end{equation}
We then construct pairwise comparison features:
\begin{equation}
    \mathbf{h}_{i,k}
    =
    \left[
    \mathbf{e}_{i,k}^{\mathrm{gt}},
    \mathbf{e}_{i,k}^{\mathrm{rec}},
    |\mathbf{e}_{i,k}^{\mathrm{gt}} - \mathbf{e}_{i,k}^{\mathrm{rec}}|,
    \mathbf{e}_{i,k}^{\mathrm{gt}} \odot \mathbf{e}_{i,k}^{\mathrm{rec}},
    \langle \mathbf{e}_{i,k}^{\mathrm{gt}}, \mathbf{e}_{i,k}^{\mathrm{rec}} \rangle
    \right].
\end{equation}
The final feature vector concatenates these features across encoders:
\begin{equation}
    \mathbf{h}_i = [\mathbf{h}_{i,1}; \mathbf{h}_{i,2}; \cdots; \mathbf{h}_{i,K}].
\end{equation}
All visual encoders are kept frozen; only the residual MLP prediction head is trained. The MLP maps $\mathbf{h}_i$ to question-level predictions in $[0,1]$. Further architecture and training details are given in Appendix Sect.~\ref{app:bcs_training_config}.

\subsection{Training Objective}
\label{sec:bcs_training}

We train BCS on the 6,855 annotated image pairs using a concept-level split to reduce leakage across semantically similar samples. The train, validation, and test proportions are 0.70, 0.15, and 0.15. The model is trained with a weighted regression loss over question targets:
\begin{equation}
    \mathcal{L}
    =
    \frac{
    \sum_{i,q} w_{i,q}
    \left(\hat{z}_{i,q} - z_{i,q}\right)^2
    }{
    \sum_{i,q} w_{i,q}
    }.
\end{equation}
The agreement weight $w_{i,q}$ gives larger weight to high-consensus VLM labels and lower weight to ambiguous labels. We compute it from the dispersion of the four VLM scores:
\begin{equation}
    w_{i,q}
    =
    1 -
    \mathrm{Var}_{v \in \mathcal{V}}
    \left(\phi(a_{i,q}^{(v)})\right).
\end{equation}
This reduces the influence of highly disputed examples while still allowing the model to learn from partially agreed annotations.

At evaluation time, we report question-level MAE, Pearson correlation, and Spearman correlation. We also compute aggregate perceptual and semantic predictions:
\begin{equation}
    \widehat{\mathrm{T\text{-}PAS}}_i
    =
    \frac{1}{|\mathcal{P}|}
    \sum_{q \in \mathcal{P}}
    \hat{z}_{i,q},
\end{equation}
\begin{equation}
    \widehat{\mathrm{T\text{-}SAS}}_i
    =
    \frac{1}{|\mathcal{S}|}
    \sum_{q \in \mathcal{S}}
    \hat{z}_{i,q},
\end{equation}
where $\mathcal{P}$ and $\mathcal{S}$ are the six perceptual and six semantic question sets. Because BCS is intended to support evaluation at the perceptual and semantic axis level, aggregate T-PAS and T-SAS prediction is the primary use case; question-level metrics are reported to assess whether the model preserves the underlying annotation structure.

\begin{table*}[t]
\centering
\caption{
Held-out performance of BCI-Coherence Score variants. 
Q-MAE, Q-$r$, and Q-$\rho$ report question-level prediction error, Pearson correlation, and Spearman correlation over individual VLM annotation questions. 
PAS and SAS report aggregate perceptual and semantic score prediction, respectively. 
Lower MAE is better; higher correlation is better.
}
\label{tab:bcs_results}
\scriptsize
\setlength{\tabcolsep}{3pt}
\renewcommand{\arraystretch}{1.08}
\begin{tabular}{@{}lccccccc@{}}
\toprule
\textbf{Model} &
\textbf{Q-MAE} &
\textbf{Q-$r$} &
\textbf{Q-$\rho$} &
\textbf{PAS MAE} &
\textbf{PAS-$r$} &
\textbf{SAS MAE} &
\textbf{SAS-$r$} \\
\midrule
SigLIP baseline
& 0.160 & 0.760 & 0.724 & 0.098 & 0.639 & 0.103 & 0.782 \\

3-VLM ordinal ensemble
& 0.118 & 0.742 & 0.739 & 0.092 & 0.632 & 0.098 & 0.786 \\

4-VLM ordinal ensemble
& 0.122 & 0.783 & 0.788 & 0.086 & 0.645 & 0.091 & 0.828 \\

4-VLM continuous ensemble
& 0.137 & 0.815 & 0.798 & 0.079 & 0.700 & 0.082 & 0.850 \\
\bottomrule
\end{tabular}
\end{table*}

\subsection{Results}
\label{sec:bcs_results}

Table~\ref{tab:bcs_results} compares several BCS variants on the held-out test split. Q-MAE, Q-$r$, and Q-$\rho$ measure question-level prediction error, Pearson correlation, and Spearman correlation over individual annotation questions, while PAS and SAS measure aggregate perceptual and semantic score prediction. The SigLIP baseline uses a single frozen visual encoder, while the ensemble variants combine multiple frozen encoders. Ordinal variants predict discretized annotation levels corresponding to the original \texttt{no}, \texttt{somewhat}, and \texttt{yes} scale, whereas the continuous variant predicts the fractional median consensus targets directly, including intermediate values such as $0.25$ and $0.75$. The 3-VLM and 4-VLM variants differ in the number of VLM teachers used to form the consensus supervision.

The four-teacher continuous ensemble gives the strongest aggregate performance, with PAS MAE of 0.079, PAS-$r$ of 0.700, SAS MAE of 0.082, and SAS-$r$ of 0.850. Although the 3-VLM ordinal ensemble obtains the lowest Q-MAE, its aggregate PAS and SAS performance is weaker than the four-teacher continuous ensemble. This shows that lower question-level error does not necessarily imply better axis-level perceptual or semantic prediction. Since BCS is intended as a reusable continuous evaluator at the perceptual--semantic axis level, we use the four-teacher continuous ensemble as the main BCI-Coherence Score.

\subsection{Analysis}
\label{sec:bcs_analysis}

The results in Table~\ref{tab:bcs_results} show why aggregate evaluation is more informative than question-level error alone. The ordinal model benefits from snapping predictions to discrete annotation levels, which improves Q-MAE, but the continuous four-teacher ensemble better preserves the fractional consensus signal produced by multiple VLMs. This leads to stronger aggregate T-PAS and T-SAS prediction, which is the intended operating point of BCS.

BCS provides three practical benefits. First, it compresses expensive multi-VLM annotation into a compact image-pair model, making BCI-aware evaluation easier to reuse across reconstruction methods. Second, it predicts both perceptual and semantic dimensions rather than collapsing reconstruction quality into a single generic similarity score. Third, its strongest performance is on aggregate semantic scoring: the four-teacher continuous ensemble reaches SAS-$r$ of 0.850, indicating that the learned evaluator captures much of the multi-VLM semantic consensus.

The difference between question-level and aggregate performance is also informative. Some individual questions, such as fine-grained color or global structure, remain difficult because they are subjective under reconstruction noise. Aggregating across perceptual or semantic dimensions produces more stable targets, which is the intended use case for BCS.

\subsection{Human Agreement}
\label{sec:human_agreement}

We use the human-rated subset from the perceptual-semantic coherence study as a targeted validation set for the perceptual-semantic regimes analyzed in this work. The subset contains 18 reconstructed images derived from 9 unique stimulus instances, with two reconstructions per stimulus. It was stratified using automated perceptual, semantic, and coherence scores: six images were selected from high-perceptual-alignment cases, six from high-semantic-alignment cases, three from the highest-coherence cases, and three from the lowest-coherence cases. This design covers perception-dominant, semantic-dominant, and coherence-extreme regimes, matching the types of disagreement that T-PAS, T-SAS, and BCS are intended to characterize.

Human evaluation was conducted with 18 annotators. Each annotator judged paired ground-truth and reconstructed images using structured perceptual, semantic, and holistic coherence questions. Perceptual questions focused on visual realism, shape/layout similarity, and texture/color consistency. Semantic questions focused on category match, key attributes such as object type or count, and whether the reconstruction depicted the same concept or scene. Coherence questions asked whether the reconstruction felt like a faithful reconstruction, whether visual appearance and semantic meaning agreed, and whether a human would accept it as representing the same object or scene.

\begin{table}[t]
\centering
\caption{
Human agreement on perceptual, semantic, and coherence judgments. 
Cohen's $\kappa$ is reported as mean $\pm$ standard deviation across annotator comparisons; Krippendorff's $\alpha$ measures overall reliability. 
Coherence judgments show substantially higher reliability than isolated perceptual or semantic judgments.
}
\label{tab:human_agreement}
\small
\setlength{\tabcolsep}{4pt}
\renewcommand{\arraystretch}{1.08}
\begin{tabular}{@{}lcc@{}}
\toprule
\textbf{Human judgment} & \textbf{Cohen's $\kappa$} & \textbf{Krippendorff's $\alpha$} \\
\midrule
Semantic agreement & $0.483 \pm 0.323$ & 0.574 \\
Perceptual agreement & $0.110 \pm 0.241$ & 0.094 \\
Coherence judgment & $0.882 \pm 0.174$ & 0.882 \\
\bottomrule
\end{tabular}
\end{table}

Table~\ref{tab:human_agreement} shows that human judgments are most reliable when perceptual and semantic evidence are evaluated jointly. Perceptual agreement alone has low reliability, with $\kappa=0.110 \pm 0.241$ and Krippendorff's $\alpha=0.094$, indicating that low-level visual fidelity is difficult to judge consistently under EEG reconstruction noise. Semantic agreement is more stable, with $\kappa=0.483 \pm 0.323$ and $\alpha=0.574$, suggesting moderate consensus on high-level content. Coherence judgments are substantially more reliable, reaching $\kappa=0.882 \pm 0.174$ and $\alpha=0.882$.

The gap between perceptual and coherence reliability directly 
motivates the joint T-PAS/T-SAS design: neither axis alone produces 
stable human judgments under EEG reconstruction noise. The validated 
subset spans perception-dominant, semantic-dominant, and 
coherence-extreme regimes, confirming that BCS targets a 
disagreement structure that human evaluators consistently recognize.

\section{Conclusion}

Standard reconstruction metrics fail systematically in EEG-to-image 
evaluation: they penalize semantically recoverable outputs and reward 
visually plausible ones that depict the wrong content. We introduced 
BCI-Coherence Score, a lightweight evaluator distilled from four-VLM 
consensus annotations over 6,855 reconstruction pairs. It separately 
predicts tolerant perceptual and semantic alignment without repeated 
VLM inference. Human validation supports the joint design: coherence 
judgments reach $\kappa = 0.882$ against $\kappa = 0.110$ for 
perceptual judgments alone. BCS provides a reusable evaluation signal 
for future EEG-to-image benchmarking; extending it to fMRI and 
cross-subject settings remains open.

\clearpage
\appendix
\renewcommand{\theHsection}{appendix.\Alph{section}}
\section{Dataset and Pair Construction}
\label{app:dataset}
The evaluated corpus contains 6,855 ground-truth/reconstruction pairs. Pair identifiers encode the reconstruction method, subject where present, concept, rank, and candidate index. BCS uses a concept-level split to reduce semantic leakage across train, validation, and held-out test partitions.

Table~\ref{tab:app_dataset} reports the composition of the evaluated reconstruction corpus. The \emph{Pairs} column gives the number of reference/reconstruction pairs contributed by each source, while \emph{Subjects} and \emph{Concepts} summarize the available diversity within that source.

\begin{table}[!htbp]
\centering
\caption{Dataset composition.}
\label{tab:app_dataset}
\small
\begin{tabular}{lrrr}
\toprule
Method & Pairs & Subjects & Concepts \\
\midrule
ATM & 1,990 & 10 & 200 \\
ENIGMA & 3,980 & 10 & 200 \\
brainvis & 200 & 1 & 200 \\
cvpr40\_brainvis & 39 & 1 & 39 \\
cvpr40\_dreamdiffusion & 46 & 1 & 46 \\
dreamdiffusion & 200 & 1 & 200 \\
thingseeg\_brainvis & 200 & 1 & 200 \\
thingseeg\_dreamdiffusion & 200 & 1 & 200 \\
\bottomrule
\end{tabular}

\end{table}

ATM and ENIGMA provide 5,970 of the 6,855 pairs and cover ten subjects and 200 concepts each. The remaining sources contribute smaller but methodologically distinct reconstruction sets, including method-specific outputs from BrainVis and DreamDiffusion and their THINGS-EEG/CVPR40 variants.

Table~\ref{tab:app_regimes} groups pairs by thresholded T-PAS and T-SAS. The four quadrants separate jointly coherent reconstructions from two disagreement regimes: high-P/low-S cases, where reconstructions appear visually plausible but depict nonmatching semantic content, and low-P/high-S cases, where semantic content remains recoverable despite weak visual fidelity.

\begin{table}[!htbp]
\centering
\scriptsize
\caption{Perceptual-semantic regimes by method using the 0.5 threshold.}
\label{tab:app_regimes}
\resizebox{\textwidth}{!}{\begin{tabular}{lrrrrr}
\toprule
Group & N & High P/High S & High P/Low S & Low P/High S & Low P/Low S \\
\midrule
overall & 6,855 & 494 (0.072) & 207 (0.030) & 455 (0.066) & 5699 (0.831) \\
ATM & 1,990 & 85 (0.043) & 75 (0.038) & 115 (0.058) & 1715 (0.862) \\
ENIGMA & 3,980 & 153 (0.038) & 121 (0.030) & 234 (0.059) & 3472 (0.872) \\
brainvis & 200 & 110 (0.550) & 5 (0.025) & 47 (0.235) & 38 (0.190) \\
cvpr40\_brainvis & 39 & 23 (0.590) & 0 (0.000) & 9 (0.231) & 7 (0.179) \\
cvpr40\_dreamdiffusion & 46 & 9 (0.196) & 0 (0.000) & 4 (0.087) & 33 (0.717) \\
dreamdiffusion & 200 & 1 (0.005) & 0 (0.000) & 1 (0.005) & 198 (0.990) \\
thingseeg\_brainvis & 200 & 112 (0.560) & 6 (0.030) & 44 (0.220) & 38 (0.190) \\
thingseeg\_dreamdiffusion & 200 & 1 (0.005) & 0 (0.000) & 1 (0.005) & 198 (0.990) \\
\bottomrule
\end{tabular}
}
\end{table}

Most pairs fall in the low-perceptual/low-semantic quadrant, reflecting the difficulty of EEG reconstruction. BrainVis outputs show more high-P/high-S cases than the larger ATM and ENIGMA sets, while DreamDiffusion outputs are mostly low on both axes in this corpus. Although less frequent, the high-P/low-S quadrant captures the semantic-blindness failure mode in which a reconstruction is visually plausible but semantically mismatched.

\FloatBarrier
\section{VLM Prompt Specification}
\label{app:prompt}
The annotation protocol is configured as a reference-grounded VLM evaluation. Each call presents the reference image and EEG reconstruction in a fixed order and applies the same six perceptual and six semantic questions. The same configuration is used for all VLM annotators.

\begin{table}[!htbp]
\centering
\scriptsize
\caption{Prompt configuration used for VLM-based annotation.}
\label{tab:app_prompt_config}
\begin{tabularx}{\textwidth}{@{}lX@{}}
\toprule
\textbf{Item} & \textbf{Configuration} \\
\midrule
Input order & Image~1 is always the ground-truth reference stimulus; Image~2 is always the EEG-generated reconstruction. The ordering is fixed across annotation, rationale generation, and score parsing. \\
Evaluation context & Reconstructions are judged under EEG-decoding limitations: blur, low resolution, missing details, distortion, low contrast, artifacts, and stylistic mismatch are expected. The prompt explicitly separates this task from generic photorealistic image-generation evaluation. \\
Question set & Both images are shown. The VLM answers six perceptual questions P1 to P6 and six semantic questions S1 to S6. \\
Answer set & Each question is answered with exactly one of \texttt{yes}, \texttt{somewhat}, or \texttt{no}. These labels are the only values admitted into scoring. \\
Rationale field & Each answer is accompanied by a short evidence sentence. Rationales are used for agreement analysis, but the scalar T-PAS/T-SAS scores use only the discrete answer labels. \\
\bottomrule
\end{tabularx}
\end{table}

Table~\ref{tab:app_prompt_config} defines the run-level annotation contract. It fixes image ordering, defines the expected reconstruction-noise context, and specifies the question and answer format. These constraints limit the tendency of VLM judges to apply generic image-generation criteria to EEG reconstructions and thereby over-penalize expected low-level degradation.

The returned annotation is parsed into question-level answer and rationale fields, grouped into perceptual and semantic blocks. The evaluator prompt text used for scoring is reproduced below.

\subsection{System Instruction}
Every annotation call used the same task instruction. The system message defined the VLM as an expert evaluator for EEG-to-image reconstruction and specified the input order: Image~1 is the ground-truth reference stimulus shown during EEG recording, and Image~2 is the EEG-generated reconstruction produced by a neural decoding model. The instruction emphasized that EEG-to-image outputs are often blurry, low resolution, missing fine detail, distorted, stylistically different from the reference, noisy, artifact-laden, low contrast, or flat in appearance. The VLM was therefore instructed to judge content preservation under EEG-specific limitations rather than photorealistic image-generation quality.

For every question, the VLM selected exactly one answer from \texttt{yes}, \texttt{somewhat}, and \texttt{no}. A \texttt{yes} answer means that the criterion is clearly satisfied, \texttt{somewhat} means that it is partially satisfied with noticeable gaps or errors, and \texttt{no} means that it is mostly not satisfied. Each answer was accompanied by a short reasoning sentence explaining the visible evidence for the label. No answers outside this three-level scale were accepted.

\subsection{Perceptual Prompts}
Each perceptual question targets one visual dimension. The VLM is instructed to answer each question independently, tolerate EEG-specific degradation such as blur, noise, and low detail, and penalize only the property specified by the question.

\paragraph{P1: Global spatial structure.}
\emph{Does the generated image preserve the coarse spatial organization of the reference, including the approximate position, size, and arrangement of the dominant regions?}
This prompt asks only where things are in the image and how the space is divided; shape, color, texture, and meaning are not considered. Tolerated deviations include blurry or softened region boundaries, approximate rather than exact positioning, simplified spatial layout, and missing background elements. Penalized errors include placing the dominant object in a completely wrong location, inverting foreground and background relationships, producing a spatial layout unrelated to the reference, or omitting the primary spatial region entirely.

\paragraph{P2: Object shape and silhouette.}
\emph{Does the dominant object or figure in the generated image have a shape or silhouette similar to the reference?}
This prompt asks only about the form of the main object: its outline, contour, and overall geometric structure. Color, texture, position, and meaning are not considered. Tolerated deviations include simplified or smoothed contours, missing fine shape detail, mild deformation of secondary parts, and blurred object boundaries. Penalized errors include a completely wrong dominant-object shape, a silhouette from a different object class, an unrecognizable or collapsed object form, or a shape that implies a different category.

\paragraph{P3: Surface texture and material.}
\emph{Does the surface texture and material appearance of the dominant object in the generated image resemble the reference?}
This prompt asks only about surface quality: smooth versus rough, matte versus shiny, organic versus manufactured, and fine-grained versus coarse. Color, shape, position, and meaning are not considered. Tolerated deviations include reduced texture resolution or fidelity, blurry or flat surface appearance, approximate rather than exact material match, and missing fine surface detail. Penalized errors include a completely wrong surface material, such as fur rendered as metal, a texture pattern belonging to a different object class, or no recoverable surface information.

\paragraph{P4: Color and chromatic consistency.}
\emph{Does the dominant color palette of the generated image reasonably match the reference?}
This prompt asks only about color: dominant hues, approximate saturation, and broad chromatic character of the main object and scene. Texture, shape, position, and meaning are not considered. Tolerated deviations include approximate rather than exact color match, reduced saturation or muted palette, slight hue shift, and missing color variation in secondary regions. Penalized errors include a completely wrong dominant color, a palette associated with a different object type, or chromatic information that is absent or inverted.

\paragraph{P5: Absence of dominant artifacts.}
\emph{Is the generated image free from severe visual artifacts that dominate or overwhelm the content?}
This prompt asks only about artifacts: structured noise, grid patterns, color bleeding, repetitive hallucinated patterns, or generation failures. Blur and low detail are not counted as artifacts. Tolerated deviations include blur, low sharpness, missing detail, flat regions, mild color noise, and soft or undefined edges. Penalized errors include structured artifacts dominating large image regions, repetitive or tiled hallucinated patterns, severe color bleeding across object boundaries, or fragmented incoherent structure from generation failure.

\paragraph{P6: Holistic visual recoverability.}
\emph{Taking the generated image as a whole, can a human observer still recover the primary visual content of the reference despite EEG reconstruction degradation?}
This is a gestalt-level prompt asking whether the overall image, across structure, shape, texture, and color together, carries enough visual evidence to identify what is being depicted. It is not answered from any single dimension alone. Tolerated deviations include any individual visual dimension being weak or missing, low overall visual quality, and degraded or noisy appearance. Penalized errors include no recoverable content across any dimension, an image that is entirely noise or artifact, or an image from which no visual content can be inferred.

\subsection{Semantic Prompts}
Semantic prompts ignore photorealism and focus on meaning. The VLM is instructed that a blurry or distorted image can still carry correct semantic content, and that each semantic question should be answered independently.

\paragraph{S1: Basic category identity.}
\emph{Does the generated image depict an object or scene belonging to the same basic-level category as the reference?}
Basic-level categories include animal, vehicle, food, furniture, tool, clothing, building, plant, person, household object, electronic device, natural scene, and indoor scene. This prompt asks only about the broadest categorical identity; specific type, attributes, count, and scene context are not considered. Example judgments include elephant to elephant as \texttt{yes}, elephant to rhinoceros as \texttt{somewhat}, elephant to car as \texttt{no}, apple to orange as \texttt{yes}, and apple to baseball as \texttt{no}. Tolerated deviations include wrong specific type within the correct category, visual degradation making details unclear, and missing secondary objects. Penalized errors include a clearly different basic-level category or categorically different scene type.

\paragraph{S2: Subordinate identity.}
\emph{Does the generated image depict the specific type of object shown in the reference, beyond just the basic category?}
This prompt asks about subordinate-level identity: not just animal but dog versus cat versus elephant, not just vehicle but car versus truck versus bicycle, and not just food but apple versus banana versus pizza. Basic category, function, count, and scene context are not considered. Example judgments include golden retriever to labrador as \texttt{yes}, golden retriever to german shepherd as \texttt{somewhat}, golden retriever to cat as \texttt{no}, sports car to sedan as \texttt{somewhat}, and sports car to truck as \texttt{no}. Tolerated deviations include approximate species or type match, genuinely ambiguous specific type, and visual degradation obscuring fine distinctions. Penalized errors include an unambiguously wrong specific type or a generated image that clearly implies a different subordinate identity.

\paragraph{S3: Functional role and purpose.}
\emph{Does the dominant object in the generated image serve the same functional role or purpose as in the reference?}
This prompt asks what the object is for or what it does, not its category or appearance. Functional roles include seating, transportation, food consumption, cutting or handling, display or communication, containment, and locomotion. Category identity, appearance, count, and scene context are not considered. Example judgments include chair to stool as \texttt{yes} because both provide seating, knife to fork as \texttt{somewhat} because both are utensils with different functions, knife to pen as \texttt{no}, car to bicycle as \texttt{somewhat} because both support transportation, and car to house as \texttt{no}. Tolerated deviations include approximate or related functional role and visual degradation obscuring object details. Penalized errors include a completely different functional role or purpose class.

\paragraph{S4: Quantity and cardinality.}
\emph{Does the generated image show approximately the same number of primary objects as the reference?}
This prompt asks only about quantity: how many main objects are present. Object identity, location, and meaning are not considered. The VLM counts only primary objects, ignores background elements, and ignores secondary or decorative objects. Example judgments include one dog to one dog as \texttt{yes}, one dog to two dogs as \texttt{no}, three apples to two apples as \texttt{somewhat}, three apples to one apple as \texttt{no}, a crowd of people to several people as \texttt{yes}, and one person to a crowd as \texttt{no}. Tolerated deviations include being off by one when the total count is large and ambiguous, missing secondary or background objects, and degradation that makes exact count genuinely unclear. Penalized errors include clearly wrong count or an unambiguous singular/plural inversion.

\paragraph{S5: Scene context and environment.}
\emph{Does the generated image imply the same scene context or environment as the reference, independent of the main object?}
This prompt asks about setting or background context, not the main object. It distinguishes scene pairs such as indoor versus outdoor, natural versus urban or manufactured, water versus land, domestic versus wild, and aerial versus ground level. Main-object identity, function, count, and appearance are not considered. Example judgments include dog in a park to dog in a field as \texttt{yes}, dog in a park to dog in a kitchen as \texttt{no}, car on highway to car on road as \texttt{yes}, car on highway to car underwater as \texttt{no}, and bird on branch to bird over trees as \texttt{yes}. Tolerated deviations include missing background detail, blurry or undefined environment, and approximate setting match. Penalized errors include categorically different context, indoor/outdoor inversion, natural/urban inversion, or a setting implying a completely different environment.

\paragraph{S6: Semantic recoverability under noise.}
\emph{Taking the generated image as a whole, can the intended semantic content of the reference be inferred from the generated image despite EEG reconstruction noise?}
This is a gestalt-level semantic prompt asking whether enough semantic evidence remains across category, identity, function, quantity, and scene to identify what the reconstruction is trying to represent. It is not answered from any single semantic dimension alone. Tolerated deviations include any individual semantic dimension being weak, low visual quality or heavy degradation, partial or approximate semantic evidence, and only one or two semantic dimensions surviving. Penalized errors include no recoverable semantic content across any dimension, a generated image that actively suggests a completely different concept, or an image from which the intended object or scene cannot be inferred.

\subsection{Answer Mapping and Scores}
All prompts use the same three-level answer scale. Let $a_{i,q}^{(v)}$ denote the parsed answer from VLM $v$ for image pair $i$ and question $q$:
\begin{equation}
\phi(\texttt{no}) = 0,\qquad
\phi(\texttt{somewhat}) = 0.5,\qquad
\phi(\texttt{yes}) = 1.
\end{equation}
Per-VLM perceptual and semantic scores are
\begin{equation}
\mathrm{T\text{-}PAS}_i^{(v)}
= \frac{1}{6}\sum_{p=1}^{6}\phi(a_{i,p}^{(v)}),\qquad
\mathrm{T\text{-}SAS}_i^{(v)}
= \frac{1}{6}\sum_{s=1}^{6}\phi(a_{i,s}^{(v)}).
\end{equation}
For downstream distillation, consensus targets are formed at the question level:
\begin{equation}
z_{i,q}=\mathrm{median}_{v\in\mathcal{V}}\phi(a_{i,q}^{(v)}),
\end{equation}
where $\mathcal{V}$ is the four-VLM annotator set. This preserves partial disagreement; with four annotators, median targets can take values such as 0.25 and 0.75. All twelve question dimensions enter the per-pair averages, BCS loss, and reported aggregate scores.

\FloatBarrier
\section{Annotation and Model Details}
\label{app:annotation_model_details}

\subsection{VLM Annotators}
\label{app:vlm_annotators}
The annotation run used four open-weight VLMs with a common prompt. The model set provides complementary calibration behavior while remaining reproducible from public checkpoints. Table~\ref{tab:app_vlm_config} lists the checkpoints and inference settings used for the annotation pass. \emph{Valid} is the fraction of image pairs that produced parseable structured annotations, while mean T-PAS and T-SAS summarize each model's operating point.

\begin{table*}[!htbp]
\centering
\scriptsize
\caption{Final VLM annotator configuration and aggregate operating points.}
\label{tab:app_vlm_config}
\resizebox{\textwidth}{!}{\begin{tabular}{lrrrrrrr}
\toprule
Annotator & Checkpoint & Dtype & Batch & Max toks & Valid & Mean T-PAS & Mean T-SAS \\
\midrule
InternVL3 & OpenGVLab/InternVL3-8B & bf16 & 1,2,4,8 & 80 & 1.000 & 0.341 & 0.252 \\
SAIL-VL & BytedanceDouyinContent/SAIL-VL-1d6-8B & bf16 & 24 & 128 & 1.000 & 0.213 & 0.157 \\
OLA-7B & THUdyh/Ola-7b & bf16 & 32 & 128 & 1.000 & 0.433 & 0.335 \\
Ovis2-8B & AIDC-AI/Ovis2-8B & bf16 & 8 & 128 & 1.000 & 0.337 & 0.345 \\
\bottomrule
\end{tabular}
}
\end{table*}

All four annotators produced valid annotations for the full corpus. Their mean scores differ meaningfully: SAIL-VL is the most conservative on both axes, OLA-7B is more permissive perceptually, and Ovis2-8B gives the highest average semantic score. This variation motivates the four-model consensus, which reduces dependence on any single model's calibration.

\subsection{Metric-to-Consensus Correlation}
\label{app:metric_consensus}
Conventional metrics are oriented so larger values indicate better reconstruction quality. Table~\ref{tab:app_metric_correlation} reports their alignment with the four-VLM consensus T-PAS and T-SAS axes. Spearman correlation measures rank agreement, while Pearson correlation measures linear association with the consensus scores.

\begin{table*}[!htbp]
\centering
\caption{Correlation between conventional metrics and VLM consensus T-PAS/T-SAS.}
\label{tab:app_metric_correlation}
\scriptsize
\begin{tabular}{lrrrr}
\toprule
Metric & $\rho$(T-PAS) & $r$(T-PAS) & $\rho$(T-SAS) & $r$(T-SAS) \\
\midrule
MSE & 0.137 & 0.124 & 0.074 & 0.056 \\
PSNR & 0.137 & 0.122 & 0.074 & 0.054 \\
SSIM & 0.208 & 0.127 & 0.060 & -0.015 \\
LPIPS & 0.144 & 0.147 & 0.239 & 0.228 \\
DISTS & 0.363 & 0.367 & 0.460 & 0.447 \\
DreamSim & 0.421 & 0.536 & 0.627 & 0.683 \\
OpenCLIP & 0.415 & 0.534 & 0.676 & 0.727 \\
DINOv2 & 0.401 & 0.571 & 0.577 & 0.701 \\
TOPIQ-FR & 0.240 & 0.244 & 0.297 & 0.296 \\
PieAPP & 0.070 & 0.073 & 0.182 & 0.168 \\
ImageReward & 0.255 & 0.488 & 0.426 & 0.625 \\
BLIP-SBERT & 0.301 & 0.409 & 0.468 & 0.554 \\
\bottomrule
\end{tabular}

\end{table*}

Low-level metrics such as MSE, PSNR, SSIM, and PieAPP align weakly with the VLM consensus, especially on T-SAS. Feature and reward metrics correlate more strongly with semantic consensus, with OpenCLIP, DreamSim, DINOv2, and ImageReward showing the clearest associations. However, these metrics still produce a single scalar similarity score and do not separate perceptual recoverability from semantic recoverability, which motivates T-PAS/T-SAS and BCS.

\subsection{BCS Training Configuration}
\label{app:bcs_training_config}
BCS distills four-VLM consensus into a compact image-pair regressor. The model is trained at the question level: for each image pair $i$ and prompt dimension $q$, the four VLM answers are first mapped to $\{0,0.5,1\}$ and aggregated into a consensus target $z_{i,q}$. Agreement weights give more influence to prompt instances where the VLMs agree and less influence to ambiguous instances.

For each frozen image encoder $E_k$, BCS computes normalized embeddings for the reference image $y_i$ and reconstruction $\hat{y}_i$:
\begin{equation}
\mathbf{e}^{\mathrm{ref}}_{i,k}=E_k(y_i),\qquad
\mathbf{e}^{\mathrm{rec}}_{i,k}=E_k(\hat{y}_i).
\end{equation}
The image-pair feature block for encoder $k$ concatenates the reference embedding, reconstruction embedding, absolute difference, elementwise product, and cosine similarity:
\begin{equation}
\mathbf{h}_{i,k} =
\left[
\mathbf{e}^{\mathrm{ref}}_{i,k};
\mathbf{e}^{\mathrm{rec}}_{i,k};
\left|\mathbf{e}^{\mathrm{ref}}_{i,k}-\mathbf{e}^{\mathrm{rec}}_{i,k}\right|;
\mathbf{e}^{\mathrm{ref}}_{i,k}\odot\mathbf{e}^{\mathrm{rec}}_{i,k};
\left\langle \mathbf{e}^{\mathrm{ref}}_{i,k},\mathbf{e}^{\mathrm{rec}}_{i,k}\right\rangle
\right].
\end{equation}
The final BCS input is the concatenation of these blocks across SigLIP, CLIP, and DINOv3, giving a 9,219-dimensional feature vector. Encoder embeddings are precomputed and cached; all encoders remain frozen during BCS training.

The prediction head is a residual MLP with a layer-normalized input projection, hidden dimension 512, two residual blocks, dropout 0.2, and sigmoid outputs for the twelve prompt dimensions. It is optimized with AdamW using learning rate $5\times10^{-4}$, weight decay $5\times10^{-4}$, batch size 512, and a maximum of 120 epochs. Training uses early stopping on validation loss with patience 16, cosine learning-rate decay, and gradient-norm clipping at 1.0. The concept-level split uses seed 42 with a 0.70/0.15/0.15 train/validation/test ratio.

For the continuous BCS model, the loss is the weighted mean-squared error over the twelve prompt dimensions:
\begin{equation}
\mathcal{L}_{\mathrm{BCS}} =
\frac{\sum_{i,q} w_{i,q}\left(\hat{z}_{i,q}-z_{i,q}\right)^2}
{\sum_{i,q} w_{i,q}},
\end{equation}
where $w_{i,q}$ is the agreement weight. The continuous ensemble averages raw predictions from five independently seeded MLPs with seeds 7, 13, 21, 37, and 42. Held-out evaluation reports question-level MAE and correlation, then aggregates predicted perceptual and semantic prompt scores into BCS estimates of T-PAS and T-SAS.

Table~\ref{tab:app_bcs_model} summarizes the implementation settings for the BCS training run. The encoder row identifies the frozen backbones, the feature-dimension row gives the concatenated pair-feature size, and the split/seed rows define the data partition and ensemble members. The optimization and loss rows summarize the training schedule used for each seed.

\begin{table*}[!htbp]
\centering
\scriptsize
\caption{BCS training and architecture details.}
\label{tab:app_bcs_model}
\resizebox{\textwidth}{!}{\begin{tabular}{l>{\raggedright\arraybackslash}p{0.74\textwidth}}
\toprule
Item & Setting \\
\midrule
Encoders & google/siglip-base-patch16-224; openai/clip-vit-large-patch14; timm/vit\_base\_patch16\_dinov3.lvd1689m \\
Feature dimension & 9,219 \\
Split & concept split, seed 42, train 0.70, val 0.15, test 0.15 \\
Seeds & 7, 13, 21, 37, 42 \\
MLP & 2 residual blocks, hidden 512, dropout 0.2 \\
Optimization & lr 0.0005, weight decay 0.0005, batch 512, epochs 120, patience 16 \\
Loss & MSE weight 1.0, CE weight 0.35 \\
Runtime & seed-7 total 27.6s; ensemble uses 5 seeds \\
\bottomrule
\end{tabular}
}
\end{table*}

\FloatBarrier

\bibliographystyle{splncs04}
\bibliography{references}

\end{document}